\documentclass{llncs}
\usepackage{amsmath}
\usepackage{amssymb}
\usepackage{graphicx}
\usepackage{algorithmicx}
\usepackage[noend]{algpseudocode}
\usepackage{algorithm}
\usepackage{color}
\usepackage{url}
\usepackage{multirow}

\newcommand{\gmat}{g}
\newcommand{\gmatv}{g^{(\lv)}}
\newcommand{\Wmatrix}{W}

\newcommand{\lapl}{L}

\newcommand{\Qvec}{Q}
\newcommand{\Svec}{\mathbf{s}}
\newcommand{\MatrixElements}[1]{\left [ #1 \right ]}

\newcommand{\funspace}{\mathcal{H}}

\newcommand{\Xset}{\mathcal{X}}
\newcommand{\Yset}{\mathcal{Y}}

\newcommand{\Tset}{\mathcal{T}}
\newcommand{\Reals}{\mathbb{R}}

\newcommand{\reg}{R}

\newcommand{\scr}{s}
\newcommand{\dat}{\mathbf{q}}
\newcommand{\ins}{\mathbf{x}}

\newcommand{\lab}{y}

\newcommand{\tsetsize}{n}
\newcommand{\tsus}{l}
\newcommand{\tsls}{n}

\newcommand{\regsize}{r}

\newcommand{\svmalpha}{a}

\newcommand{\lv}{v}
\newcommand{\lu}{u}
\newcommand{\lM}{M}

\newcommand{\kernelf}{k}
\newcommand{\learningf}{f}
\newcommand{\costfunction}{c}

\newcommand{\disagrarg}{d}
\newcommand{\objective}{J}

\newcommand{\transpose}{^{t}}
\newcommand{\ttr}{t}
\newcommand{\inv}{^{-1}}
\newcommand{\Prefx}{\mathcal{P}_{\ins}}
\newcommand{\PrefFx}{\mathcal{P}_{f,\ins}}

\newcommand{\OComplex}{\mathcal{O}}
\newcommand{\sign}{\textrm{sign}}

\newcommand{\dict}{\mathcal{D}}
\newcommand{\res}{\mathbf{r}}

\newcommand{\Po}{N}

\newcommand{\kk}{\mathbf{k}}
\newcommand{\gk}{\bar{\mathbf{k}}}
\newcommand{\fkk}{\mathbf{f}}

\newcommand{\Kmat}{K}

\newcommand{\kmatRR}{\Kmat_{R,R}}

\newtheorem{fct}{Proposition}

\begin{document}

\title{Semi-supervised Ranking Pursuit}
\author{Evgeni Tsivtsivadze\inst{1} \and Tom Heskes\inst{2}}
\institute{
The Netherlands Organization for Applied Scientific Research, \\ Zeist, The Netherlands \\
\email{evgeni.tsivtsivadze@tno.nl},
\and
Institute for Computing and Information Sciences, \\ Radboud University Nijmegen, The Netherlands\\
\email{t.heskes@science.ru.nl}}

\maketitle
\thispagestyle{empty}

\begin{abstract}
We propose a novel sparse preference learning/ranking algorithm. Our algorithm
approximates the true utility function by a weighted sum of basis functions
using the squared loss on pairs of data points, and is a generalization of the
kernel matching pursuit method. It can operate both in a supervised and a semi-supervised
setting and allows efficient search for multiple, near-optimal solutions.
Furthermore, we describe the extension of the algorithm suitable for combined ranking and regression tasks.  
In our experiments we demonstrate that the proposed algorithm outperforms several
state-of-the-art learning methods when taking into account unlabeled data and performs
comparably in a supervised learning scenario, while providing sparser solutions.
\end{abstract}

\section{Introduction}

Recently, preference learning \cite{plbook} has received
significant attention in machine learning community. Informally, the main goal of this task is prediction
of ordering of the data points rather than prediction of a numerical score as in the case of regression
or a class label as in the case of a classification task. The ranking problem
can be considered as a special case of preference learning when a strict
order is defined over all data points.
The applications of algorithms that learn to rank\footnote{See e.g. http://research.microsoft.com/en-us/um/beijing/projects/letor/paper.aspx} are widespread, including
information retrieval (collaborative filtering, web search e.g.\ \cite{cortes2007magnitude}),
natural language processing (parse ranking e.g.\ \cite{collins_pr}), bioinformatics (protein ranking e.g.\ \cite{weston_rank}), and many others.

Despite notable progress in the development and application of preference
learning/ranking algorithms (e.g.\ \cite{plbook}), so far the emphasis was
mainly on improving the learning performance of the methods. 
Much less is known about models that focus
in addition on interpretability and sparseness of the ranking solution. In this work we 
propose a novel preference learning/ranking algorithm that besides state-of-the-art
performance can also lead to sparse models   
and notably faster prediction times (that is an absolute necessity for a wide range of
applications such as e.g.\ search engines), compared to the non-sparse
counterparts.

Sparse modeling is a rapidly developing area of machine learning motivated by 
the statistical problem of variable selection in high-dimensional datasets. The aim is to
obtain a highly predictive (small) set of variables that can help to
enhance our understanding of underlying phenomena. This objective 
constitutes a crucial difference between sparse modeling and other
machine learning approaches.
Recent developments in theory and algorithms for sparse modeling mainly concern
l1-regularization and convex relaxation of the subset selection problem. Examples of such algorithms
include sparse regression (e.g.\ Lasso \cite{lasso}) and its various extensions (Elastic Net \cite{Zou05regularizationand}, 
group Lasso \cite{Yuan06modelselection,Bach07consistencyof}, simultaneous/multi-task Lasso \cite{adaptiveXing}) as well as sparse dimensionality reduction
(sparse PCA \cite{adaptiveXing}, NMF \cite{LeeS00}) algorithms. Applications of these methods are wide-ranging, 
including computational biology, neuroscience, image processing,
information retrieval, and social network analysis, to name a few. 
The \textit{sparse} ranking algorithm we propose here is not tied to a particular domain 
and can be applied to various problems 
where it is necessary to estimate preference relations/ranking of the objects 
as well as to obtain a compact and representative model.

RankSVM \cite{Joachims} is a ranking method that can lead to sparse solutions.
However, in RankSVM sparsity control is not
explicit and the produced models are usually far from being interpretable.
Also note that frequently ranking algorithms are not directly applicable
to more general preference learning tasks or can become computationally expensive.
Our method is a generalization of the (kernel) matching pursuit algorithm \cite{kmp}
and it approximates the true utility function by a weighted sum of basis functions
using squared loss on pairs of data points.
Unlike existing methods our algorithm allows explicit control over sparsity
of the model and can be applied to ranking and preference learning problems.

Furthermore, an extension of the algorithm allows us to 
efficiently search for several near-optimal solutions instead of a single one.
For example, some of the problems that arise during the sparse modeling include 
possible existence of multiple nearly-optimal solutions (e.g.\ due to the lack of a single sparse
ground truth). This situation is common for  many biological problems when,
for example, finding a few highly predictive proteins does not exclude the possibility 
of finding some other group of genes/proteins with similar properties.
The same situation can occur in many other domains, e.g. information retrieval 
(various groups of highly descriptive document/queries in document ranking task),
natural language processing (parse re-ranking), etc.
Therefore, it is an important issue to explore and include search for multiple
nearly-optimal sparse solutions rather than a single solution.
In our empirical evaluation we show that our algorithm can operate
in supervised and semi-supervised settings, leads to sparse solutions, and improved performance compared
to several baseline methods.
  
\section{Problem Setting}
\label{sec:problem_setting}
Let $\Xset$ be a set of instances and $\Yset$ be a set of labels. 
We consider the \emph{label ranking} task \cite{plbook,DekelMS03} namely, 
we want to predict for any instance $ \ins \in \Xset$  a preference relation
$\Prefx \subseteq \Yset\times\Yset$ among the set of labels $\Yset$.
We assume that the true preference relation $\Prefx$ is transitive and
asymmetric for each instance $\ins \in \Xset$. Our training set
${\{(\dat_i,\scr_i)\}}_{i=1}^\tsetsize$ contains the data points
$(\dat_i,\scr_i)=((\ins_i,\lab_i),\scr_i) \in \left(\Xset\times\Yset\right) \times \Reals$
that are an instance-label tuple $\dat_i=(\ins_i,\lab_i) \in \Xset\times\Yset$
and its score $\scr_i \in \Reals$. We define the pair of data points $((\ins,\lab),\scr)$ 
and $((\ins',\lab'),\scr')$ to be  \emph{relevant}, iff $\ins=\ins'$ and \emph{irrelevant} otherwise.

As an example, consider an information retrieval task where every query
is associated with the set of retrieved documents. The intersection of the
retrieved documents associated with different queries can be either empty or
non-empty. We are usually interested in ranking the documents that are
associated with a single query (the one that has retrieved the documents).
Thus, ranks between documents retrieved by different queries are
\emph{not relevant} for this task, whereas those documents retrieved
by the same query are \emph{relevant}.

Given a relevant pair $((\ins,\lab),\scr)$ and $((\ins,\lab'),\scr')$, we
say that instance $\ins$ \emph{prefers} label $\lab$ to $\lab'$, iff $\scr > \scr'$.
If $\scr=\scr'$, the labels are called \emph{tied}. Accordingly, we write
$\lab \succ_{\ins} \lab'$ if $\scr > \scr'$ and $\lab \sim_{\ins} \lab'$ if $\scr=\scr'$.
Finally, we define our training set $\Tset=(\Qvec,\Svec,\Wmatrix)$, where
$\Qvec=(\dat_1,\ldots,\dat_\tsetsize)\transpose\in {\left( \Xset \times \Yset\right)}^\tsetsize$ 
is the vector of instance-label training tuples and $\Svec=(\scr_1,\ldots,\scr_\tsetsize)\transpose\in\Reals^\tsetsize$
is the corresponding vector of scores. The $\Wmatrix$ matrix defines a preference
graph and incorporates information about relevance of a particular data point to the task, e.g.\ 
$\MatrixElements{\Wmatrix}_{i,j}=1$, if  $(\dat_i,\dat_j), 1 \le i,j \le n, i \ne j,$
are relevant and 0 otherwise.

Informally, the goal of our ranking task is to find a \emph{label ranking function}
such that the ranking $\PrefFx \subseteq \Yset\times\Yset$ induced by the function
for any instance $\ins \in \Xset$ is a good ``prediction'' of the true preference
relation $\Prefx \subseteq \Yset \times \Yset$. Formally, we search for the function
$\learningf : \Xset \times \Yset \rightarrow \Reals$ mapping each instance-label
tuple $(\ins,\lab)$ to a real value representing the (predicted) relevance of the
label $\lab$ with respect to the instance $\ins$. To measure how well a hypothesis
$ \learningf$ is able to predict the preference relations $\Prefx$ for all
instances $\ins \in \Xset$, we consider the following cost function
(disagreement error) that captures the amount of incorrectly predicted pairs of relevant training data points:
\begin{equation}\label{diserrorsignum}
\disagrarg(\learningf,\Tset)
= \frac{1}{2} \sum_{i,j=1}^\tsetsize \MatrixElements{\Wmatrix}_{i,j}
\Big\arrowvert\sign\big( \scr_i-\scr_j \big)
-\sign\big( \learningf(\dat_i)-\learningf(\dat_j)\big)\Big\arrowvert,
\end{equation}
where $\sign (\cdot)$ denotes the signum function.

\section{Ranking Pursuit}
\label{rp}
In this section we tailor the kernel matching pursuit algorithm \cite{kmp}
to the specific setting of preference learning and ranking.
Considering the training set $\Tset=(\Qvec,\Svec,\Wmatrix)$ and a dictionary
of functions  $\dict =\{ \kernelf_1, \ldots, \kernelf_\Po \}$,
where $\Po$ is the number of functions in the dictionary. 
We are interested in finding a sparse approximation of the  
prediction function $\learningf_P(\dat) = \sum_{p=1}^P 
\svmalpha_p \kernelf_{\gamma_{p}}(\dat)$ using the basis functions 
$ \{ \kernelf_1, \ldots, \kernelf_P \} \subset \dict $ and the 
coefficients $ \{ \svmalpha_1, \ldots, \svmalpha_P \}  \in \Reals^{P}$.
The order of the dictionary functions as they appear in the expansion 
is given by a set of indices $\{\gamma_1,\ldots,\gamma_P\}$, where
$\gamma \in \{1,\ldots,N\}$. Basis functions  
are chosen to be kernel functions, similar to \cite{kmp}, that is
$\kernelf_{\gamma}(\dat) = \kernelf(\dat_{\gamma}, \dat)$ with $\kernelf(\cdot,\cdot)$
an appropriate kernel function. We will use the notation $\fkk_P = (\learningf_P(\dat_1), \ldots, 
\learningf_P(\dat_\tsetsize) )\transpose$ to represent the $\tsetsize$-dimensional 
vector that corresponds
to the evaluation of the function on the training points and similarly
$\kk_\gamma = (\kernelf_{\gamma}(\dat_1), \ldots, \kernelf_{\gamma}(\dat_n))\transpose$.
We also define $\res = \Svec - \fkk_P$  to be the residue.
The basis functions and the corresponding coefficients are to 
be chosen such that they minimize an approximation of the disagreement error:
\begin{figure*}
\begin{center}
\setlength\fboxsep{5.0pt}
\setlength\fboxrule{0.3pt}
\fbox{\includegraphics[width=0.95\columnwidth]{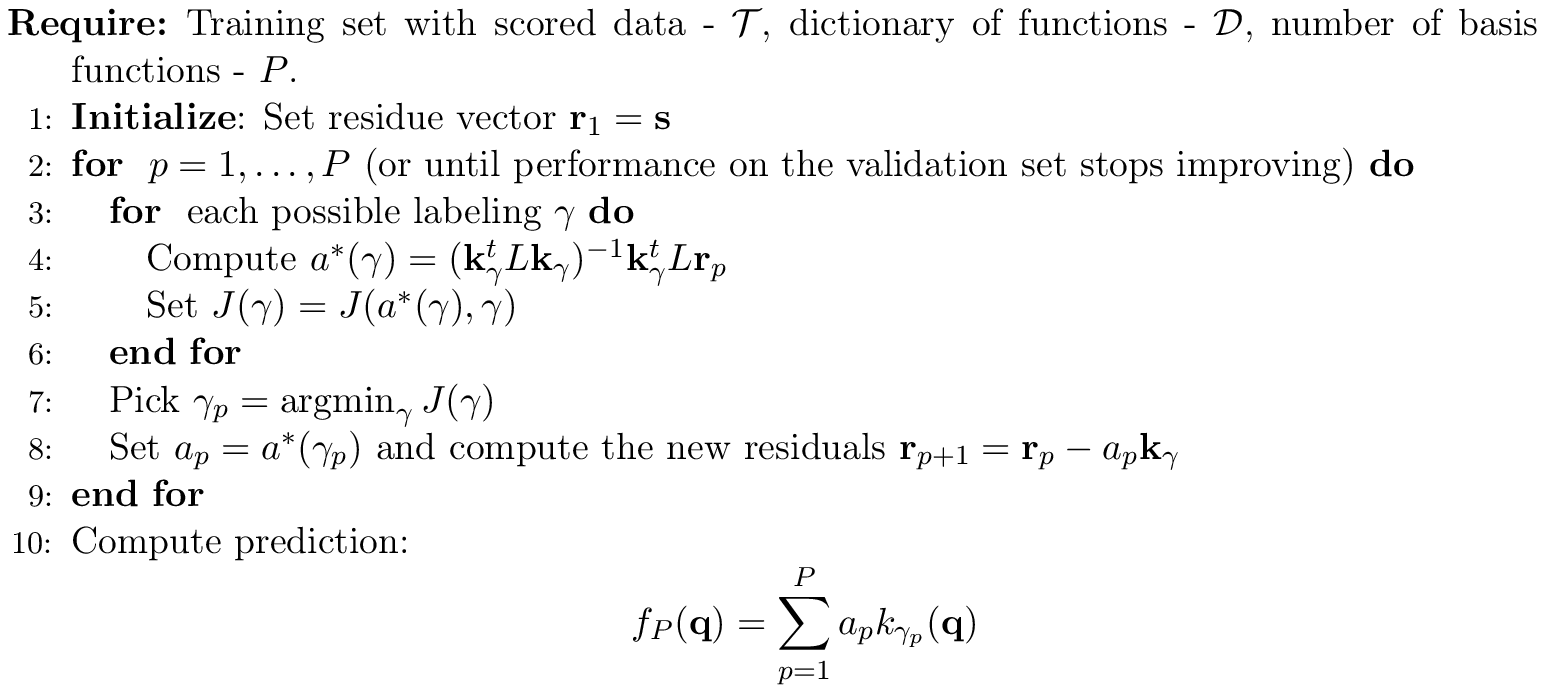}}
\caption{Supervised ranking pursuit algorithm.}
\label{fig:originallll}
\end{center}
\end{figure*}

\begin{equation*}
\costfunction(\learningf_{P},\Tset)
=\frac{1}{2} \sum_{i,j=1}^\tsetsize \MatrixElements{\Wmatrix}_{i,j}
{\Big((\scr_i - \scr_j) - (\learningf_{P}(\dat_i) - \learningf_{P}(\dat_j))\Big)}^2,
\end{equation*}
which in matrix form can be written as
\begin{equation}
\costfunction(\fkk_{P},\Tset)
= (\Svec - \fkk_{P})\transpose \lapl(\Svec - \fkk_{P}),
\end{equation}
where $\lapl$ is the Laplacian matrix of the graph $\Wmatrix$.

\newcommand{\istage}{p}

The ranking pursuit starts at stage 0, with $\fkk_{0}$, and recursively
appends functions to an initially empty basis, at each stage
of training to reduce the approximation of the 
ranking error. Given $\fkk_{\istage}$  we build
$
\fkk_{\istage +1 } (\svmalpha, \gamma)
= \fkk_{\istage} + \svmalpha \kk_{\gamma},
$ by searching for $ \gamma \in \{1,\ldots,N \}$ and
$\svmalpha \in \Reals$ such that at every step (the residue of) the error is minimized:   

\begin{eqnarray*}
\objective(\svmalpha, \gamma) &=& 
\costfunction(\fkk_{p+1}(\svmalpha, \gamma),\Tset) = (\Svec - \fkk_{\istage+1}(\svmalpha, \gamma)) \transpose
 \lapl(\Svec - \fkk_{\istage+1}(\svmalpha, \gamma))\\ &=& (\res_\istage - \svmalpha \kk_{\gamma}) 
\transpose \lapl( \res_\istage - \svmalpha \kk_{\gamma}),
\end{eqnarray*}
where we define $\kernelf_{\gamma i}= \kernelf(\dat_\gamma, \dat_i)$ 
and $\kk_\gamma = (\kernelf_{\gamma 1}, \ldots, \kernelf_{\gamma n} )\transpose$. 
By setting the first derivative to zero and solving the resulting system of equations we can obtain 
the coefficient $\svmalpha$ that minimizes $\objective(\svmalpha, \gamma)$ for a 
given $\gamma$, that is  $\svmalpha = 
(\kk_\gamma\transpose\lapl\kk_\gamma)\inv\kk_\gamma\transpose\lapl\res_p$.

The set of basis functions and coefficients obtained
at every iteration of the algorithm is suboptimal. This can be
corrected by a back-fitting procedure using a least-squares approximation
of the disagreement error. The optimal value of the parameter $P$,
that can be considered a ``regularization'' parameter of the algorithm,
is estimated using a cross-validation procedure. The pseudo-code for
the algorithm is presented in Figure 1.

\subsection{Learning Multiple Near-Optimal Solutions}

In this subsection we formulate an extension of the ranking pursuit
algorithm that can efficiently use unscored data to improve 
the performance of the algorithm. The main idea behind our approach
is to construct multiple, near-optimal, ``sparse'' ranking functions
that give a small error on the scored data and whose
predictions agree on the unscored part. 

Semi-supervised learning algorithms have gained more and more attention in recent years
as unlabeled data is typically much easier to obtain than labeled data.
\emph{Multi-view} learning algorithms, such as co-training \cite{blum},
split the attributes into independent sets and an algorithm is learnt
based on these different ``views''. 
The goal of the learning process consists of finding a prediction
function for every view (for the learning task) that performs
well on the labeled data of the designated view such that all
prediction functions agree on the unlabeled data. Closely related to
this approach is the \emph{co-regularization} framework described in \cite{co-Reg,tsivtsivadze10b},
where the same idea of agreement maximization between the predictors is central.
Briefly stated, algorithms based upon this approach search for hypotheses
from different Reproducing Kernel Hilbert Spaces (RKHS) \cite{schoelkopf2001representer},
namely views, such that the training error of each hypothesis on the labeled
data is small and, at the same time, the hypotheses give similar predictions
for the unlabeled data. Within this framework, the disagreement is takens
into account via a co-regularization term. Empirical results show that
the co-regularization approach works well for classification \cite{co-Reg},
regression \cite{effco-Reg}, and clustering tasks \cite{ICML2011Kumar_272,mv2011nips}. Moreover,
theoretical investigations demonstrate that the co-regularization approach
reduces the Rademacher complexity by an amount that depends
on the ``distance'' between the views \cite{randemaxer07,sindhwaniR08}.

Let us consider $\lM$ different feature spaces $\funspace_1,\ldots,$ $\funspace_\lM$
that can be constructed from different data point descriptions (i.e.,
different features) or by using different kernel functions. 
Similar to \cite{kmp} we consider $\funspace$ to be a RKHS.
In addition to the training set $\Tset=(\Qvec,\Svec,\Wmatrix)$
originating from a set ${\{(\dat_i,\scr_i)\}}_{i=1}^\tsls$ of data
points \emph{with} scoring information. We also have a training set
$\overline{\Tset}=(\overline{\Qvec},\overline{\Wmatrix})$ from a
set ${\{\overline{\dat}_{i}\}}_{i=1}^\tsus$ of data points 
\emph{without} scoring information, $\overline{\Qvec} = (\overline{\dat}_{1},
\ldots, \overline{\dat}_{\tsus})\transpose \in 
{\left( \Xset \times \Yset\right)}^\tsus$, and the corresponding
adjacency matrix $\overline{\Wmatrix}$. To avoid misunderstandings
with the definition of the label ranking task, we will use the terms
``scored'' instead of ``labeled'' and ``unscored'' instead of ``unlabeled''.
We search for the functions $F_P = (\learningf_P^{(1)},\ldots,
\learningf_P^{(\lM)}) \in \funspace_1 \times \ldots \times\funspace_\lM$, minimizing
\begin{equation}\label{eq:coregloss_unlabelled}
\widetilde{\costfunction}(F_P,\Tset, \overline{\Tset}) =
\sum_{\lv=1}^\lM \costfunction(\learningf^{(\lv)}_P, \Tset) 
+\nu \sum_{\lv,\lu=1}^\lM \overline{\costfunction}
(\learningf^{(\lv)}_P,\learningf^{(\lu)}_P, \overline{\Tset}),
\end{equation}
where $ \nu \in \Reals^+$ is a regularization parameter and where
$\overline{\costfunction}$ is the loss function measuring the
disagreement between the prediction functions of the views on the unscored data:

\begin{eqnarray*}
 \overline{\costfunction} (\learningf^{(\lv)}_P, \learningf^{(\lu)}_P, \overline{\Tset}) &&=
 \frac{1}{2} \sum_{i,j=1}^\tsus {\MatrixElements{\overline{\Wmatrix}}}_{i,j}
 \Big(\big( \learningf^{(\lv)}_P (\overline{\dat}_{i}) - \learningf^{(\lv)}_P (\overline{\dat}_{j}) \big) - \\
  &&\big( \learningf^{(\lu)}_P (\overline{\dat}_{i}) - \learningf^{(\lu)}_P (\overline{\dat}_{j})\big)
\Big)^2 \notag.
\end{eqnarray*}
Although we have used unscored data in our formulation, we note that the
algorithm can also operate in a purely supervised setting.
It will then not only minimize the error on the scored data but 
also enforce agreement among the prediction functions constructed from different views.

The prediction functions
$\learningf^{(\lv)}_P \in \funspace_\lv$ of (\ref{eq:coregloss_unlabelled})
for $\lv =1,\ldots,\lM$ have the form
$ \learningf^{(\lv)}_P(\dat) = \sum_{p=1}^{P} \svmalpha_p^{(\lv)} \kernelf_{\gamma_v p}^{(\lv)}(\dat)$ 
with corresponding coefficients $\{ \svmalpha_1^{(\lv)},\ldots,\svmalpha_{P}^{(\lv)} \}\in \Reals^P$. 
Let $\bar{\lapl}$  denote the Laplacian matrix of the graph $\overline{\Wmatrix}$.
Using a similar approach as in section 3 we can write the objective function as 

\begin{eqnarray*}
\lefteqn{ \objective(\mathbf{\svmalpha},\boldsymbol\gamma) =
\widetilde{\costfunction}(F_{p+1}(\mathbf{\svmalpha},\boldsymbol\gamma),\Tset,\overline{\Tset}) =} \\ 
&&\sum_{\lv=1}^\lM (\res_p - \svmalpha^{(\lv)} \kk^{(\lv)}_{\gamma_v})\transpose
\lapl(\res_p - \svmalpha^{(\lv)} \kk^{(\lv)}_{\gamma_v}) + \\ 
&&\nu\sum_{\lv,\lu=1}^\lM  ( \svmalpha^{(\lv)} \gk^{(\lv)}_{\gamma_v} - \svmalpha^{(\lu)}
\gk^{(\lu)}_{\gamma_{u}})\transpose \bar{\lapl} (\svmalpha^{(\lv)} \gk^{(\lv)}_{\gamma_{v}} -
\svmalpha^{(\lu)} \gk^{(\lu)}_{\gamma_{u}}),
\end{eqnarray*}
where $\mathbf{\svmalpha} = (\svmalpha^{(1)},\ldots, \svmalpha^{(M)} )\transpose \in \Reals^M$,
$\boldsymbol\gamma =  (\gamma_{1},\ldots, \gamma_{M})$ with $\gamma_{\lv} \in \{1,\ldots,N\}$,
and $\gk_\gamma$ is the basis vector expansion on unscored data with $\bar{\kernelf}_{\gamma i}=
\kernelf(\overline{\dat}_\gamma, \overline{\dat}_i)$. 
By taking partial derivatives with respect to the coefficients in each view (for clarity
we denote $\kk^{(\lv)}_{\gamma_{v}}$ and $\gk^{(\lv)}_{\gamma_{v}}$ as $\kk^{(\lv)} \text{and }
\gk^{(\lv)}$, respectively) and 
defining $\gmatv_\nu = 2\nu(\lM -1)\gk^{(\lv)\ttr}  \bar{\lapl} \gk^{(\lv)}$ 
and $\gmatv = \kk^{(\lv)\ttr} \lapl \kk^{(\lv)}$, we obtain

\begin{eqnarray*}
\frac{\partial}{\partial \svmalpha^{(\lv)}} \objective({\mathbf{\svmalpha}}, \boldsymbol\gamma)
= 2(\gmatv + \gmatv_\nu)\svmalpha^{(\lv)} -2 \kk^{(\lv)\ttr}  \lapl \res_p
- 4\nu\sum_{\lu=1, \lu \ne \lv }^\lM \gk^{(\lv)\ttr} \bar{\lapl} \gk^{(\lu)} \svmalpha^{(\lu)}. 
\end{eqnarray*}
At the optimum we have $\frac{\partial}{\partial \svmalpha^{(\lv)}} \objective({\mathbf{\svmalpha}},\boldsymbol\gamma)=0$ 
for all views, thus,  we get the exact solution by solving

\begin{equation*}
\left(
\begin{array}{ccc}
\gmat^{(1)} + \gmat^{(1)}_{\nu} & -2\nu \gk^{(1)\ttr}  \bar{\lapl} \gk^{(2)} & \ldots \\
\\
-2\nu  \gk^{(2)}  \bar{\lapl} \gk^{(1)} & \gmat^{(2)} + \gmat^{(2)}_{\nu} & \ldots \\
\\
\vdots & \vdots & \ddots
\end{array} \right)
\left(
\begin{array}{cc}
\svmalpha^{(1)} \\
\\
\svmalpha^{(2)} \\
\\
\vdots
\end{array} \right)
= 
\left(
\begin{array}{cc}
\kk^{(1)\ttr}\lapl\res_p \\
\\
\kk^{(2)\ttr}\lapl\res_p \\
\\
\vdots 
\end{array} \right)
\end{equation*}
with respect to the coefficients in each view. 
Note that the left-hand side matrix is positive definite by construction and,
therefore, invertible. Once the coefficients are estimated, multiple solutions
can be obtained using the prediction functions constructed for each view.

We can also consider a single prediction function that is given, for example, by the
average of the functions for all views. The overall complexity of the standard ranking
pursuit algorithm is $\OComplex(P n^2)$, thus, there is no increase in computational
time compared to the kernel matching pursuit algorithm in the supervised setting \cite{kmp}.
The semi-supervised version of the ranking pursuit algorithm requires $\OComplex(P n^M (\lM^3 + \lM^2 l))$ time,
which is linear in the number of unscored data points\footnote{In semi-supervised
learning usually $n \ll l$, thus, linear complexity in the number of unscored data points is
beneficial. We note that complexity of the algorithm can be further reduced 
to $\OComplex(PM^3nl)$ by forcing the indices of the nonzero coefficients in the different views to be the same.}.
The pseudo-code for the algorithm is presented in Figure 2. 
\begin{figure*}
\begin{center}
\setlength\fboxsep{5.0pt}
\setlength\fboxrule{0.3pt}
\fbox{\includegraphics[width=0.95\columnwidth]{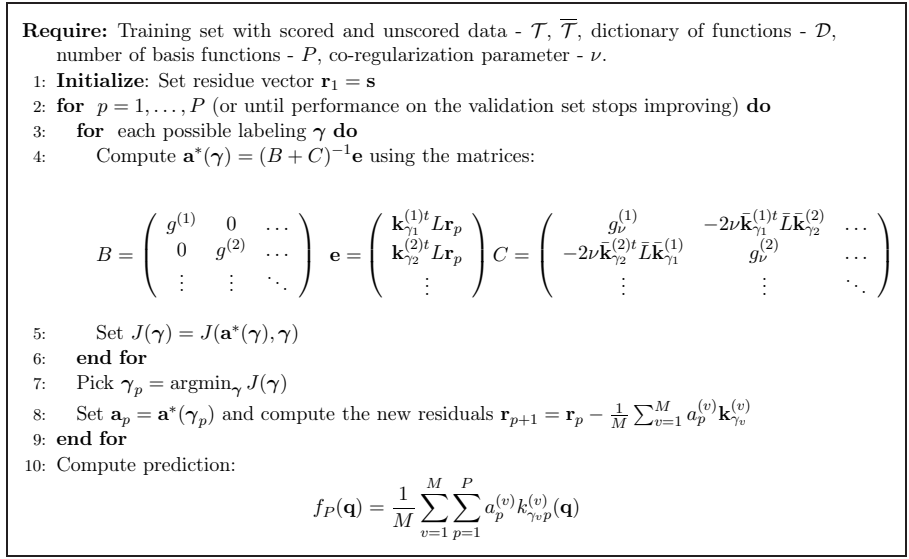}}
\caption{Semi-supervised ranking pursuit algorithm.}
\label{fig:originallll}
\end{center}
\end{figure*}

\section{Combined Ranking and Regression Pursuit}
\label{rrp}

Recently a method on combined ranking and regression has been proposed in \cite{google}.
The authors suggest that in many circumstances it is beneficial to minimize the combined objective  
function simultaneously due to the fact that the algorithm can avoid learning degenerate
models suited only for some particular set of performance metrics. 
Furthermore, such objective can help to improve regression performance
in some circumstances e.g.\ when there is a large class imbalance situation. Empirically, 
the combined approach gives the ``best of both''
performance, performing as well at regression
as a regression-only method, and as well at ranking as a ranking-only only method.
However, despite the efficient stochastic gradient descent algorithm described in \cite{google}
the objective function to be minimized still consists of two separate parts, namely regression and
ranking with the appropriate weight coefficients attached to both. 

Motivated by the above approach and strong empirical results presented in \cite{google} we
propose a framework for joint ranking and regression optimization based on our ranking pursuit algorithm. 
We argue that our approach is slightly more elegant and simpler compared to \cite{google} due to the fact that
we employ a generalization of kernel matching pursuit -- a genuine regression algorithm,
thus, we do not have to consider two separate objective functions when learning joint ranking and regression
models.

Compared to the kernel matching pursuit algorithm which minimizes least-squares error function 
\begin{equation*}\label{costmp}
\costfunction(\learningf_{P},\Tset)
=\frac{1}{2} \sum_{i=1}^\tsetsize
{\Big(\scr_i - \learningf_{P}(\dat_i)\Big)}^2,
\end{equation*}
recall that the supervised ranking pursuit chooses the basis functions and the corresponding coefficients
such that they minimize an approximation of the disagreement error:
\begin{equation*}\label{ccrank}
\costfunction(\learningf_{P},\Tset)
=\frac{1}{2} \sum_{i,j=1}^\tsetsize \MatrixElements{\Wmatrix}_{i,j}
{\Big((\scr_i - \scr_j) - (\learningf_{P}(\dat_i) - \learningf_{P}(\dat_j))\Big)}^2,
\end{equation*}
which in matrix form can be written as
\begin{equation} \label{crr}
\costfunction(\fkk_{P},\Tset)
= (\Svec - \fkk_{P})\transpose \lapl(\Svec - \fkk_{P}),
\end{equation}
where $\lapl$ is the Laplacian matrix of the graph $\Wmatrix$ defined in section \ref{sec:problem_setting}. Note that
we can obtain a standard regression algorithm by using an identity matrix instead of $\lapl$ in (\ref{crr}).
A simple idea behind our combined ranking and regression approach is the appropriate selection of the weights for 
the matrix $\lapl$, so that in a special case we can obtain a regression formulation and in another we can
recover complete pairwise ranking. For this purpose we consider the weighted Laplacian matrix
\begin{equation}
\tilde{\lapl} = \beta I + (1-\beta)\lapl.
\end{equation}
By setting the $\beta$ coefficient equal to zero, we recover the standard ranking pursuit.
On the other hand by setting $\beta$ equal to 1 we obtain kernel matching pursuit \cite{kmp}.
By setting the values of the coefficient
between these extremes and using such weighted $\tilde{\lapl}$ in (\ref{crr}) corresponds
to minimizing a ``combined'' ranking and regression objective function.
We refer to this algorithm as combined ranking and regression pursuit (CRRP).

\section{Subset of Regressors Method for Ranking Algorithms}
\label{randproj}

For comparison with the state-of-the-art, we will compare our algorithm with the sparse RankRLS, 
recently proposed in \cite{pahikkala2009preferences}.  
The main idea behind sparse RankRLS is to employ 
subset selection method (described e.g.\ in \cite{GPbook}) that is generally used for 
computational speed up purposes.
For example, a popular approach to speed up the algorithm consists
in approximating the kernel matrix. However, this in turn leads to solutions that do not depend on
all data points present in the training set and, thus, can be considered as sparse.

Let us briefly describe this approach:
Consider a setup when instead of selecting a basis function to minimize the disagreement error
at every iteration of the algorithm as in section \ref{rp},
we choose the prediction function to have the following form:
$\learningf(\dat) = \sum_{p=1}^\tsetsize \svmalpha_p \kernelf(\dat,\dat_p)$. Further, 
given the prediction function that depends on all training data the objective
function in matrix form can be written as $(\Svec-K\mathbf{a})\transpose L (\Svec-K\mathbf{a})$, where 
$ \Kmat \in \Reals^{\tsetsize \times \tsetsize}$ is a kernel matrix
constructed from the training set and $ \mathbf{a} = (\svmalpha_1,\ldots,\svmalpha_\tsetsize )\transpose \in \Reals^\tsetsize$
is a corresponding coefficient vector.

Now, let $\reg=\{i_1,\ldots,i_\regsize\} \subseteq [\tsetsize]$ be a subset of indices such that only
$\svmalpha_{i_1},\ldots,\svmalpha_{i_\regsize}$ are nonzero. 
By randomly selecting a subset of data points, we can approximate the prediction function using
$\hat{\learningf} (\dat) = \sum_{j=1}^r \svmalpha_{p_j} \kernelf(\dat,\dat_{p_j})$. Similarly we 
can approximate the kernel matrix and define matrix $\kmatRR \in \Reals^{r \times r}$ 
that contains both rows and columns indexed by $\reg$. 
This approach for matrix approximation, known as ``subset of regressors'', was pioneered 
in \cite{poggio} and is frequently applied in practice. 
Although it may seem over-simplistic (e.g.\ other methods might appear to be more suitable rather than
random selection of the regressors) it is efficient and usually leads to quite  
good performance. The reason behind this is that the solution obtained using a subset
of regressors method can be shown to be equivalent to a ``non-sparse'' solution
obtained with some other kernel function (e.g.\ \cite{pahikkala2009preferences}). 

In our experiments we evaluate the performance of the sparse RankRLS algorithm
and compare it to the supervised and semi-supervised ranking pursuit algorithms. 
We demonstrate that selection of the non-zero coefficients based on iterative
minimization of the disagreement error (strategy used by the ranking pursuit algorithm)
leads to better results compared to random subset selection.

\section{Experiments}
\label{experimentssection}

\subsection{Jester joke dataset}

We perform a set of experiments on the publicly available Jester joke dataset\footnote{Available at
\url{http://www.ieor.berkeley.edu/~goldberg/jester-data/}.}. 
The task we address is the prediction of the joke preferences of a user based on the 
preferences of other users. The dataset contains 4.1M ratings in the range from $-10.0$ to $+10.0$ of 100 jokes assigned by a group of 
73421 users.
Our experimental setup is similar to that of~\cite{cortes2007magnitude}. We have grouped the users into three groups according to the 
number of jokes they have rated: $20-40$ jokes, $40-60$ jokes, and $60-80$ jokes. 
The test users are randomly selected among the users who had rated between 
$50$ and $300$ jokes.
For each test user half of the preferences is reserved for training and half for testing. The preferences are derived from the differences of the ratings the test user gives to jokes, e.g.\ a joke with
higher score is preferred over the joke with lower score. The features for each test user are generated as follows.
A set of 300 reference users is selected at random from one of the three groups and their ratings for the corresponding jokes are used as a feature values. In case a user has not rated the joke the median of his/her ratings is used as the feature value.
The experiment is done for 300 different test users and the average performance is recorded. Finally, we repeat the complete experiment ten times 
with a different set of 300 test users selected at random. We report the average value over the ten runs for each of the three groups.

In this experiment we compare performance of the ranking pursuit algorithm to several algorithms, namely
kernel matching pursuit \cite{kmp}, RankSVM \cite{Joachims}, RLS \cite{rifkin-reg} (also known as kernel ridge regression \cite{Saunders98}, proximal-svm \cite{fm:01}, ls-svm \cite{suykens2002lssvm}), RankRLS and sparse RankRLS \cite{pahikkala2009preferences} in terms of the disagreement error (\ref{diserrorsignum}).
\begin{table}[t]
 \vskip 0.1in
\caption{ 
Performance comparison of the learning algorithms in supervised experiment conducted
on Jester joke dataset. Normalized version of the disagreement error is used as a performance evaluation measure. 
Note that despite performance similar to that of ranking algorithms, ranking pursuit leads on average to 30\% 
sparser solutions.
}\label{table:datasets}
\centering
\vskip 0.3in

\begin{sc}
\resizebox{0.55\textwidth}{!}{
\begin{tabular}{lccc}
\hline
Method &  $\mathbf{20-40}$ &  $\mathbf{40-60}$ &  $\mathbf{60-80}$   \\[0.01in]
\hline
RLS 		 & 0.425 & 0.419 & 0.383    \\[0.01in]
Matching Pursuit & 0.428 & 0.417 & 0.381   \\[0.01in]
RankSVM 	 & 0.412 & 0.404 & 0.372   \\[0.01in]
RankRLS 	 & 0.409 & 0.407 & 0.374   \\[0.01in]
Sparse RankRLS 	    & 0.414 & 0.410 & 0.380   \\[0.01in]
Ranking Pursuit  & 0.410 & 0.404 & 0.373   \\[0.01in]
\hline
\end{tabular}
}
\end{sc}
  \vskip 0.1in
\end{table}

\begin{table}[t]
 \vskip 0.1in
\caption{ 
Performance comparison of the learning algorithms in semi-supervised experiment
conducted on Jester joke dataset. Supervised learning methods are trained only on the scored part of the dataset.
Normalized version of the disagreement error is used as a performance evaluation measure. 
Note that semi-supervised ranking pursuit notably outperforms other methods.}
\label{table:ssmdatasets}
\centering
\vskip 0.3in
\begin{sc}
\resizebox{0.55\textwidth}{!}{
\begin{tabular}{lccc}
\hline
Method &  $\mathbf{20-40}$ &  $\mathbf{40-60}$ &  $\mathbf{60-80}$  \\[0.01in]
\hline
RLS		    & 0.449 & 0.434 & 0.405   \\[0.01in]
Matching Pursuit    & 0.451 & 0.433 & 0.404  \\[0.01in]
RankSVM 	    & 0.428 & 0.417 & 0.391   \\[0.01in]
RankRLS 	    & 0.429 & 0.418 & 0.393   \\[0.01in]
Sparse RankRLS 	    & 0.431 & 0.424 & 0.397   \\[0.01in]
Ranking Pursuit	    & 0.428 & 0.417 & 0.393   \\[0.01in]
SS Ranking Pursuit  & \textbf{0.419} & \textbf{0.411} & \textbf{0.381}   \\[0.01in]
\hline
\end{tabular}
}
\end{sc}
  \vskip 0.1in
\end{table}
In all algorithms we use a Gaussian kernel where the width 
parameter is chosen from the set $\{ 2^{-15},2^{-14}\ldots,2^{14},2^{15}\}$ and other parameters (e.g.\ stopping criteria)
are chosen by taking the average over the performances on a hold out-set.
The hold-out set is created similarly as the corresponding training/test set.

The results of the collaborative filtering experiment are included in
Table~\ref{table:datasets}. It can be observed that ranking based approaches in general outperform the regression methods. 
According to Wilcoxon signed-rank test \cite{Demsar2006} the differences in performance are 
statistically significant ($p < 0.05$). However, the differences in performance among the ranking/regression algorithms
are not statistically significant. 
Although performance of the ranking pursuit algorithm is similar to that of the RankSVM and RankRLS algorithms, obtained
solutions are on average $30\%$ sparser. 
To evaluate the performance of the semi-supervised extension of the ranking pursuit algorithm
we construct datasets similarly as in the supervised learning
experiment with the following modification. To simulate unscored data, for each test user we make only half of his/her preferences from the training set  available for learning.
Using this training set we construct two views, each containing half of the scored and half of the unscored data points. 
The rest of the experimental setup follows the previously described supervised learning setting.
The results of this experiment are included in Table~\ref{table:ssmdatasets}.
We observe notable improvement in performance of the semi-supervised ranking pursuit algorithm compared to
all baseline methods. This improvement is statistically significant according to Wilcoxon signed-rank test
with 0.05 as a significance threshold. The performance of the supervised methods in this 
experiment is decreased (compared to the supervised learning experiment) as expected,
due to the fact that the amount of labeled data is twice smaller.

\begin{table}[t]
 \vskip 0.1in
\caption{ 
Performance comparison of the learning algorithms in supervised experiment conducted
on MovieLens dataset. Normalized version of the disagreement error is used as
a performance evaluation measure. Note that despite performance similar to that of ranking algorithms, 
ranking pursuit leads on average to 35\% sparser solutions.}\label{table:datasetsm1}
\centering
\vskip 0.3in
\begin{sc}
\resizebox{0.55\textwidth}{!}{
\begin{tabular}{lccc}

\hline
Method &  $\mathbf{20-40}$ &  $\mathbf{40-60}$ &  $\mathbf{60-80}$   \\[0.01in]
\hline
RLS		    & 0.495 & 0.494 & 0.482   \\[0.01in]
Matching Pursuit    & 0.494 & 0.497 & 0.484  \\[0.01in]
RankSVM 	    & 0.481 & 0.472 & 0.453   \\[0.01in]
RankRLS 	    & 0.479 & 0.472 & 0.455   \\[0.01in]
Sparse RankRLS 	    & 0.484 & 0.478 & 0.460   \\[0.01in]
Ranking Pursuit	    & 0.480 & 0.472 & 0.453   \\[0.01in]
\hline
\end{tabular}
}
\end{sc}
  \vskip 0.1in
\end{table}
\begin{table}[t]
 \vskip 0.1in
\caption{ 
Performance comparison of the learning algorithms in semi-supervised experiment
conducted on MovieLens dataset. Supervised learning methods are trained only on the scored part of the dataset.
Normalized version of the disagreement error is used as a performance evaluation measure. Note that semi-supervised ranking pursuit
notably outperforms other methods.}
\label{table:ssmdatasetsm2}
\centering
\vskip 0.3in
\begin{sc}
\resizebox{0.55\textwidth}{!}{
\begin{tabular}{lccc}
\hline
Method &  $\mathbf{20-40}$ &  $\mathbf{40-60}$ &  $\mathbf{60-80}$   \\[0.01in]
\hline
RLS		    & 0.497 & 0.495 & 0.487   \\[0.01in]
Matching Pursuit    & 0.498 & 0.495 & 0.485  \\[0.01in]
RankSVM 	    & 0.487 & 0.479 & 0.464   \\[0.01in]
RankRLS 	    & 0.486 & 0.479 & 0.463   \\[0.01in]
Sparse RankRLS 	    & 0.490 & 0.485 & 0.470   \\[0.01in]
Ranking Pursuit	    & 0.487 & 0.479 & 0.462   \\[0.01in]
SS Ranking Pursuit  & \textbf{0.481} & \textbf{0.474} & \textbf{0.458}   \\[0.01in]
\hline
\end{tabular}
}
\end{sc}
  \vskip 0.1in
\end{table}

\subsection{MovieLens dataset}

The MovieLens dataset consists of approximately 1M  
ratings by 6,040 users for 3,900 movies. Ratings are integers from 1 to 5.  
The experiments were set-up in the same way as for the Jester joke dataset.
The results of the supervised experiment are presented in Table~\ref{table:datasetsm1}. 
Similarly to the results obtained on Jester joke dataset we observe that 
the ranking pursuit algorithm leads to much more compact models, about 
$35\%$ sparser, while having performance comparable to that of the ranking algorithms.
The results of the semi-supervised experiment are presented in Table~\ref{table:ssmdatasetsm2}. 
We again observe notable improvement in performance of the semi-supervised ranking pursuit algorithm compared to
all baseline methods. The improvement is statistically significant ($p < 0.05$). 

\subsection{CRRP Algorithm Evaluation}

To empirically evaluate our approach, termed combined ranking and regression pursuit (CRRP), we conduct experiments using the CRRP
algorithm on the Jester jokes dataset. The experiments are conducted following the supervised learning setup described above. 
We use the disagreement error and the mean squared error (MSE) to measure performance of the algorithm
for ranking and regression setting, respectively. The obtained results are presented in Table \ref{crr_table}.
\begin{table}[h!]
\label{crr_table}
 \vskip 0.1in
\caption{ 
Performance comparison of the ranking, regression, and CRRP algorithms on the Jester joke dataset.
For the performance evaluation in regression task we use
mean squared error (MSE) and for the ranking task we use a normalized
version of the disagreement error. Note that CRRP is able to achieve
good performance in both ranking and regression settings.
}\label{crr_table}
\centering
\vskip 0.3in
\begin{sc}
\resizebox{0.55\textwidth}{!}{
\begin{tabular}{lccc}

\hline 
Ranking task &  $\mathbf{20-40}$ &  $\mathbf{40-60}$ &  $\mathbf{60-80}$  \\[0.01in]
\hline 
RLS 		 & 0.425 & 0.419 & 0.383  \\[0.01in] 
Matching Pursuit & 0.428 & 0.417 & 0.381  \\[0.01in] 
RankSVM 	 & 0.412 & 0.404 & 0.372  \\[0.01in] 
RankRLS 	 & 0.409 & 0.407 & 0.374  \\[0.01in] 
Sparse RankRLS 	    & 0.414 & 0.410 & 0.380  \\[0.01in] 
Ranking Pursuit  & 0.410 & 0.404 & 0.373  \\[0.01in] 
CRRP    & \textit{0.413} & \textit{0.408} & \textit{0.373}  \\[0.01in] 

\hline
Regression task &  $\mathbf{20-40}$ &  $\mathbf{40-60}$ &  $\mathbf{60-80}$ \\[0.01in]
\hline
RLS 		 & 21.6 & 19.3 & 15.2   \\[0.01in]
Matching Pursuit & 20.1 & 18.7 & 14.9 \\[0.01in]
RankSVM 	 & 34.2 & 31.6 & 28.9  \\[0.01in]
RankRLS 	 & 33.8 & 31.3 & 29.2  \\[0.01in]
Sparse RankRLS 	 & 36.5 & 33.8 & 32.9  \\[0.01in]
Ranking Pursuit  & 34.0 & 30.9 & 29.1  \\[0.01in]
CRRP    & \textit{23.2} & \textit{20.4} & \textit{17.3}  \\[0.01in]

\hline
\end{tabular}
}
\end{sc}
  \vskip 0.1in
\end{table}
It can be observed that by choosing the weight coefficient appropriately ($\beta$ = 0.5)
the CRRP algorithm performs almost as well as the specialized algorithms on ranking and 
regression tasks. 
Note that when considering the regression setting the CRRP algorithm improves over the MSE performance of the rank-only
methods, but does not outperform the regression only methods.
Furthermore, the performance differences of the CRRP algorithm to the ranking methods on the regression task as well as 
to the regression methods on the ranking task are statistically significant according a Wilcoxon signed-rank test ($p < 0.05$).
To summarize, while CRRP does not outperform specialized algorithms in regression or ranking, it 
is able to achieve notably better performance compared to the regression-only methods for the ranking task or
ranking-only methods for the regression task.

\section{Conclusions}
We propose sparse preference learning/ranking algorithm as well as its semi-supervised extension.
Our algorithm is a generalization of the kernel matching pursuit algorithm \cite{kmp} and 
allows explicit control over sparsity of the solution. It is also naturally applicable in circumstances
when one is interested in obtaining multiple near-optimal solutions that frequently arise during
the sparse modeling of many problems in biology, information retrieval, natural language processing, etc.
Another contribution of this paper is a combined ranking and regression (CRRP) method, formulated within
the framework of the proposed ranking pursuit algorithm.

The empirical evaluation demonstrates that
in the supervised setting our algorithm outperforms regression methods
such as kernel matching pursuit, RLS and performs comparably to the RankRLS, sparse RankRLS and RankSVM
algorithms, while having sparser solutions. In its semi-supervised setting our ranking pursuit algorithm notably outperforms
all baseline methods. We also show that CRRP algorithm is suitable for learning combined ranking and regression 
objectives and leads to good performance in both ranking and regression tasks. 
In the future we aim to apply our algorithm in other domains and will examine 
different aggregation techniques for multiple sparse solutions.

\bibliography{myBibliography}
\bibliographystyle{splncs}

\end{document}